\pdfoutput=1
\documentclass[letterpaper]{article}

\usepackage[utf8]{inputenc} 
\usepackage{url}            
\usepackage{booktabs}       
\usepackage{amsfonts}       
\usepackage{amsmath, amsthm}        
\usepackage{nicefrac}       
\usepackage{microtype}      
\usepackage{xcolor}         
\usepackage{graphicx}       
\usepackage{subcaption}     
\usepackage{tabularx, tabulary}
\usepackage{tikz}           
\usepackage{scalerel}       
\usepackage{nameref}        
\usepackage{siunitx}        
\usepackage{float}          

\usepackage[]{aaai23}  
\usepackage{times}  
\usepackage{helvet}  
\usepackage{courier}  
\usepackage{url}  
\usepackage{graphicx} 
\usepackage{natbib}  
\usepackage{caption} 
\usepackage{algorithm}
\usepackage{algorithmic}

\usepackage{newfloat}
\usepackage{listings}
\DeclareCaptionStyle{ruled}{labelfont=normalfont,labelsep=colon,strut=off} 
\floatstyle{ruled}
\newfloat{listing}{tb}{lst}{}
\floatname{listing}{Listing}
\newtheorem{definition}{Definition}
\newtheorem{assumption}{Assumption}
\newtheorem{theorem}{Theorem}

\definecolor{backcolour}{rgb}{0.95,0.95,0.92}

\usetikzlibrary{positioning}

\newcommand{\reals}{{\mathbb R}}

\newcommand{\st}{\mbox{s.t.}}

\newcommand\norm[1]{\left\lVert#1\right\rVert}
\newcommand{\diag}[1]{\mbox{\textbf{diag}}(#1)}

\newcommand{\eg}{{\it e.g.}}
\newcommand{\ie}{{\it i.e.}}

\title{State-driven Implicit Modeling for Sparsity and Robustness in Neural Networks}

\author {
    Alicia Y. Tsai\textsuperscript{\rm 1}\thanks{Corresponding author},
    Juliette Decugis\textsuperscript{\rm 2},
    Laurent El Ghaoui\textsuperscript{\rm 1},
    Alper Atamt{\"u}rk\textsuperscript{\rm 3}
}
\affiliations {
    \textsuperscript{\rm 1} Department of Electrical Engineering and Computer Science, UC Berkeley \\
    \textsuperscript{\rm 2} Department of Mathematics, UC Berkeley \\
    \textsuperscript{\rm 3} Department of Industrial Engineering and Operations Research, UC Berkeley \\
    \{aliciatsai, jdecugis, elghaoui, atamturk\}@berkeley.edu
}

\begin{document}
\maketitle

\begin{abstract}
Implicit models are a general class of learning models that forgo the hierarchical layer structure typical in neural networks and instead define the internal states based on an ``equilibrium'' equation, offering competitive performance and reduced memory consumption. However, training such models usually relies on expensive implicit differentiation for backward propagation. In this work, we present a new approach to training implicit models, called State-driven Implicit Modeling (SIM), where we constrain the internal states and outputs to match that of a baseline model, circumventing costly backward computations. The training problem becomes convex by construction and can be solved in a parallel fashion, thanks to its decomposable structure. We demonstrate how the SIM approach can be applied to significantly improve sparsity (parameter reduction) and robustness of baseline models trained on FashionMNIST and CIFAR-100 datasets.
\end{abstract}

\section{Introduction}
Conventional neural networks are built upon a hierarchical architecture, where input information is processed through several recursive layers \cite{Goodfellow-et-al-2016}. Canonical examples of this include standard feed-forward networks or convolutional networks used to perform image classification \cite{10.5555/2999134.2999257,simonyan2015deep}. Recent work has proposed a more general perspective, where the internal states are implicitly defined through an ``equilibrium'' equation~\cite{NEURIPS2019_01386bd6, NEURIPS2018_69386f6b, doi:10.1137/20M1358517}, allowing for loops in the model's computational graph. As illustrated in \citeauthor{NEURIPS2020_3812f9a5, 10.5555/3495724.3496729}, implicitly-defined models are able to match state-of-the-art performance of explicitly-defined models on several tasks. In fact, the implicit framework is a more general model with greater capacity to possibly model novel architectures and prediction rules for deep learning that are not necessarily tied to any notion of ``layers.''

The forward pass of an implicit model usually relies on solving an  algebraic equation using methods such as fixed-point equations \cite{doi:10.1137/20M1358517}, ODE solvers \cite{dupont2019augmented}, or root-finding methods \cite{NEURIPS2020_3812f9a5}. The backward pass involves differentiating through the implicit equation, which usually relies on expensive black-box solvers, projected gradient descent or approximate gradients \cite{geng2021training}. The costly backward computation remains a challenge in training and evaluation of implicit models. In this work, we develop a novel method to circumvent computing the backward pass. We start from a baseline model (\eg, a pre-trained layered neural network) and constrain the states and outputs of the implicit model to match those baseline states. 
The SIM training problem is strictly feasible and convex by construction, and thus can be solved efficiently, bypassing the expensive implicit differentiation. Additionally, the method is very scalable: it can be implemented in parallel provided that the objective is decomposable across its internal state, which is usually the case. 


We find that with our approach, the number of training samples required to efficiently and effectively train an implicit model is significantly reduced. For example, using 20\%-30\% of total training data is enough to train an implicit model on CIFAR-100 dataset. Our method can also be combined with additional objectives such as sparsity (parameter reduction) or improving robustness, making it a versatile training scheme. 

Our main contributions are summarized as follows. First, we introduce the general \textbf{S}tate-driven \textbf{I}mplicit \textbf{M}odeling (SIM) training scheme to efficiently learn an implicit model by matching the internal state and outputs of a baseline model. In addition, we present simple ways to obtain the internal state and outputs from either a standard (layered) neural network or an implicit model. Third, we demonstrate how to apply SIM for parameter reduction and robustness. Our experimental results display a competitive performance of our method on both FashionMNIST and CIFAR-100 datasets. Finally, we study the visualization of the trained models and observe interesting properties of implicit models, motivating future directions of research.




\section{Related Work}

\paragraph{Implicit Models.}
Recent works \citep{NEURIPS2019_01386bd6, NEURIPS2018_69386f6b, doi:10.1137/20M1358517, Winston_Kolter_2020} have proposed an emerging ``implicitly-defined'' structure in deep learning, where the intermediate hidden states are defined via a ``equilibrium'' (fixed-point) equation, and the outputs are determined only implicitly by the equilibrium solution of such underlying equilibrium equation. Researchers have developed different classes of implicit models and demonstrated their potential in graph neural networks \cite{10.5555/3495724.3496729}, differential equation models \cite{NEURIPS2018_69386f6b}, physical control \cite{NEURIPS2018_842424a1}, and many others \cite{NEURIPS2018_ba6d843e}. \citet{pruning_deq} show that these implicitly-defined models can be successfully pruned, reducing their training and inference complexity.

\paragraph{Sparsity.} 
The growing size and computational cost of deep learning have motivated the property of sparsity, in order to reduce the size of networks by selectively zeroing out unnecessary model parameters (\textit{pruning}). This leads to a more efficient model that operates in the same high-dimensional feature space, but with a reduced representational complexity. One of the popular approaches is to remove parameters with the smallest magnitude, a technique called \textit{magnitude pruning} \cite{NIPS2015_ae0eb3ee, DBLP:journals/corr/HanMD15, 80236, DBLP:conf/iclr/ZhuG18, DBLP:conf/iclr/MolchanovTKAK17}. Magnitude pruning eliminates weights based on a learned magnitude or criterion of parameters with an \textit{a priori} threshold \cite{DBLP:journals/corr/abs-1912-08881}, which requires trial-and-error or heuristics. Others have considered a more principled way of determining the importance of parameters, including \textit{structured pruning} \cite{NEURIPS2021_ce6babd0, Chen2021OnlyTO} and \textit{directional pruning} \cite{10.5555/3495724.3496897}. Recent works have also explored the problem using optimization techniques, such as \textit{convex pruning} \cite{Aghasi2020FastCP} or the \textit{perspective reformulation technique} in  \cite{Frangioni2006PerspectiveCF, perps-screen}. 

\paragraph{Robustness.} Starting with \citeauthor{Szegedy2014}, a large number of works have shown that state-of-the-art deep neural networks (DNNs) are vulnerable to adversarial samples \cite{GoodfellowSS14, KurakinGB17a, PapernotMJFCS16}. The vulnerability of DNNs has motivated the study of building models that are robust to such perturbations \cite{madry2018towards, PapernotM0JS16, RaghunathanSL18, Gowal2018}. Defense strategies against adversarial examples have primarily focused on training with adversarial examples \cite{arxiv.1705.07204, madry2018towards} or with a carefully designed penalty loss \cite{NEURIPS2019_0defd533, GuneetStochasticPruning}.

\section{Preliminaries}
\paragraph{Notations.}
Throughout the paper, we use $n, m, p, q$ to denote the number of internal states, the number of input samples, the dimension of input vectors, and the dimension of output vectors, respectively. For a matrix $V$, $|V|$ denotes its absolute value (\textit{i.e.} $|V|_{ij} = |V_{ij}|$); $\norm{V}_0$ is its cardinality, \textit{i.e.,} the number of non-zero entries of $V$; $\norm{V}_{\infty}$ is the max-row-sum matrix operator norm; $\norm{V}_F$ is the Frobenious norm. Finally,  $\lambda_{\text{pf}}(M)$ denotes the \textit{Perron-Frobenius (PF) eigenvalue} of a square non-negative matrix $M$ \cite{doi:10.1137/1.9781611971262}.

\begin{assumption}[component-wise non-expansive]
\label{assum:cone-map}
A function $\phi$ is \textbf{co}mponent-wise \textbf{n}on-\textbf{e}xpansive (CONE) if
\[
\forall \; u, v \in \reals^n ~:~ |\phi(u) - \phi(v)| \le |u - v|,
\]
with inequality and absolute value taken component-wise.
\end{assumption}

\begin{figure}[!h]
    \centering
    \scalebox{.45}{
    \begin{tikzpicture}[
        node distance=1.5cm,
        every node/.style={fill=white, font=\sffamily, 
                           inner sep=4mm}, 
        base/.style = {rectangle, draw=black,
                       minimum width=2cm, minimum height=1cm,
                       text centered, font=\sffamily, very thick},
        align=center]
        \node (weights) [base] {$\scaleto{\begin{bmatrix}A & B \\C & D\end{bmatrix}}{65pt}$};
        \node (activation) [base, above=of weights]  {$\scaleto{\phi}{20pt}$};
        \node (input) [yshift=-3mm, xshift=1cm, right=of weights] {$\scaleto{U}{15pt}$};
        \node (output) [yshift=-3mm, left=of weights] {$\scaleto{\hat{Y} = CX + DU}{20pt}$};
        \node (post-activation state) [above=of input] {$\scaleto{X = \phi(Z)}{20pt}$};
        \node (pre-activation state) [xshift=1cm, above=of output] {$\scaleto{Z = AX + BU}{15pt}$};
        \draw[ultra thick, ->] (input.west) -- ([yshift=-3mm] weights.east);
        \draw[ultra thick, ->] ([yshift=-3mm] weights.west) -- (output.east);
        \draw[ultra thick, -] ([yshift=5mm] weights.west) -| (pre-activation state.south);
        \draw[ultra thick, ->] (pre-activation state.north) |- (activation.west);
        \draw[ultra thick, -] (activation.east) -| (post-activation state.north);
        \draw[ultra thick, ->] (post-activation state.south) |- ([yshift=5mm] weights.east);
    \end{tikzpicture}
    }
    \caption{A block-diagram view of an implicit model, where $Z$ is the pre-activation state ``before'' passing through the activation function $\phi$ and $X$ is the post-activation state ``after'' passing through $\phi$.}
    \label{fig:blockdiag}
\end{figure}
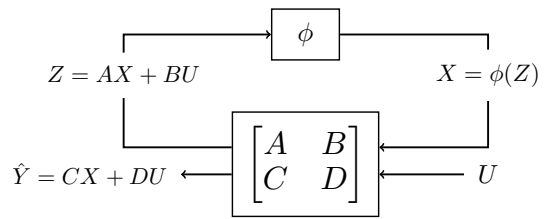

We are given a data set with input matrix $U \in \reals^{p \times m}$ and output matrix $Y \in \reals^{q \times m}$, where each column represents an input or output vector.
An implicit model consists of an equilibrium equation in a ``state matrix'' $X \in \reals^{n \times m}$ and a prediction equation:
\begin{subequations}
\label{eq:implicit-form}
\begin{align}
    &X = \phi(AX + BU) \quad \mbox{(equilibrium equation)} \label{eq:eq-eq} \\
    &\hat{Y}(U) = CX + DU \quad \mbox{(prediction equation)} \label{eq:pred-rule}
\end{align}
\end{subequations}
where $\phi:\reals^{n \times m} \rightarrow \reals^{n \times m}$ is a nonlinear activation that is strictly increasing and satisfies Assumption (\ref{assum:cone-map}), such as ReLU, tanh, or sigmoid.  While the above model seems very specific, it covers as special cases most known architectures arising in deep learning. Matrices $A\in\reals^{n\times n}$, $B\in\reals^{n\times p}$, $C\in\reals^{q\times n}$ and $D\in\reals^{q\times p}$ are model parameters. In equation (\ref{eq:eq-eq}), the input feature matrix $U \in \reals^{p \times m}$ is passed through a linear transformation by weight matrix $B$ and the internal state matrix $X \in \reals^{n \times m}$ is obtained as the fixed-point solution to equation (\ref{eq:eq-eq}). The output prediction $\hat{Y}$ is then obtained by feeding the state $X$ through the prediction equation (\ref{eq:pred-rule}). The structure is illustrated in Figure \ref{fig:blockdiag}, where the ``pre-activation'' and ``post-activation'' state matrices $Z,X$ are shown; in those matrices, each column corresponds to a single data point.

The forward pass of an implicit model relies on the fixed-point solution of the underlying equilibrium equation, while a backward pass requires one to differentiate this equation with respect to the model parameters $(A,B,C,D)$. The solution to the equilibrium equation (\ref{eq:eq-eq}) does not necessarily exists nor be unique. We say that an equilibrium equation with activation map $\phi$ is well-posed if the following \textit{well-posedness} condition is satisfied \cite{doi:10.1137/20M1358517}.
\begin{definition}[well-posedness]
The $n \times n$ matrix $A$ is said to be well-posed for $\phi$ if, for any $b \in \reals^{n}$, the solution $x \in \reals^{n}$ of the following equation
\begin{equation}
\label{def:wp}
    x = \phi(Ax+b) 
\end{equation}
exists and is unique.
\end{definition}

\paragraph{Scaling the network.}
Consider a standard layer-based neural network $\mathcal{N}: \reals \times \reals \rightarrow \reals \times \reals$ with activation $\phi$ that satisfies Assumption (\ref{assum:cone-map}) and maps input feature matrix $U \in \reals^{p \times m}$ to outputs $Y=\mathcal{N}(U)$ via hidden layers. As shown in \citeauthor{doi:10.1137/20M1358517}, for such networks, there exists an equivalent implicit model, $(A_{\mathcal{N}}, B_{\mathcal{N}}, C_{\mathcal{N}}, D_{\mathcal{N}}, \phi)$ as in (\ref{eq:implicit-form}). Without loss of generality, we may re-scale the original weight matrices of $\mathcal{N}$ to obtain a strongly well-posed implicit model, $(A'_{\mathcal{N}}, B'_{\mathcal{N}}, C'_{\mathcal{N}}, D'_{\mathcal{N}}, \phi)$, by Theorem (\ref{thm:rescaled}), in the sense that $\norm{A'_{\mathcal{N}}}_{\infty} < 1$. This result also allows us to consider the convex constraint $\norm{A}_{\infty} < 1$ as a sufficient condition as opposed to the non-convex PF sufficient condition, in light of the bound $\lambda_{\text{pf}}(A) \le \norm{A}_{\infty}$.

\begin{theorem}[PF sufficient condition for well-posedness]
\label{thm:pf}
Assume that $\phi$ satisfies Assumption (\ref{assum:cone-map}). Then, $A$ is well-posed for any such $\phi$ if $\lambda_{\text{pf}}(|A|) < 1$. Moreover, the solution $x$ of equation ($\ref{eq:eq-eq}$) can be computed via the fixed point iterations $x \rightarrow \phi(Ax+b)$, with initial condition $x = 0$.
\end{theorem}

\begin{theorem}[Rescaled implicit model]
\label{thm:rescaled}
Assume that $\phi$ is CONE and positively homogeneous, \textit{i.e.,} $\phi(\alpha x) = \alpha \phi(x)$ for any $\alpha \ge 0$ and $x$. For a neural network $\mathcal{N}$ with its equivalent implicit form $(A_{\mathcal{N}}, B_{\mathcal{N}}, C_{\mathcal{N}}, D_{\mathcal{N}}, \phi)$, where $A_{\mathcal{N}}$ satisfies PF sufficient condition for well-posedness of Theorem (\ref{thm:pf}), there exists a linearly-rescaled equivalent implicit model $(A'_{\mathcal{N}}, B'_{\mathcal{N}}, C'_{\mathcal{N}}, D'_{\mathcal{N}}, \phi)$ with $\norm{A'_{\mathcal{N}}}_{\infty} < 1$ that gives the same output $\hat{Y}$ as the original $\mathcal{N}$ for any input $U$.
\end{theorem}

The proofs of Theorem (\ref{thm:pf}) and (\ref{thm:rescaled}) are given in the appendices.
\section{State-driven Implicit Modeling}
The \textbf{S}tate-driven \textbf{I}mplicit \textbf{M}odeling (SIM) framework trains an implicit model with a constraint: it should match both the state $X$ and outputs $\hat{Y}$ of another ``baseline'' (implicit or layered) model, when the same inputs $U$ are applied. For a given baseline model, the state matrix $X$ can be obtained by running a set of fixed-point iterations (if the baseline is implicit), or a simple forward pass (if the baseline is a standard layered network). In both cases, we can extract the pre-activation state matrix $Z$, such that the post-activation state matrix satisfies $X = \phi(Z)$. Each column of matrices $Z$ and $X$ corresponds to a single data point; when the baseline is a layered network, these matrices are constructed by stacking all the intermediate layers into a long column vector, where the first intermediate layer is at the bottom and the last intermediate layer is on top.

We give a simple example of how to construct $X, Z$ from a 3-layer fully-connected network of the form:
\[
 \hat{y}(u) = W_2 x_2, \; x_2 = \phi(W_1 x_1) \; x_1 = \phi(W_0 x_0), \; x_0 = u,
\]
where $u$ is a single vector input. For notational simplicity, we exclude the bias terms, which can be easily accounted for by considering the vector $(u, 1)$ instead of $u$. Each column of $Z$ and $X$ corresponds to the state from a single input. The column $z$ is formed by stacking all the intermediate layers before passing through $\phi$ and the column $x$ is formed by stacking all the intermediate layers after passing through $\phi$:
\[
z = \begin{pmatrix} W_1 x_1 \\ W_0 x_0 \end{pmatrix}, \;\; x = \phi(z) =  \begin{pmatrix} x_2 \\ x_1 \end{pmatrix}.
\]
In this example, we can easily verify that its equivalent implicit from is as follows:
\[
\renewcommand\arraystretch{1.3}
\left( \begin{array}{c|c} A & B \\ \hline C & D \end{array} \right)= \left( \begin{array}{cc|c}0 & W_1 & 0 \\ 0 & 0 & W_0 \\ \hline W_2 & 0 & 0 \end{array} \right).
\]
For a more complicated network, finding an equivalent implicit form may be a non-trivial task. The SIM framework allows us to consider any baseline deep neural networks without ever having to address this challenge: we simply need to extract the pre- and post-activation state matrices. 

With matrices $X,Z$ fixed, we now consider the training problem, where the model parameters are encapsulated in a partitioned matrix $M \in \reals^{(n+q) \times (n+p)}$ and we define $\Tilde{Y}, \Tilde{U}$ as follows:
\[
    M := \begin{pmatrix}
    A & B \\ C & D
    \end{pmatrix}, \; 
    \Tilde{Y} := \begin{pmatrix} Z \\ \hat{Y} \end{pmatrix}, \;
    \Tilde{U} := \begin{pmatrix} X \\ U \end{pmatrix}
\]
The condition $\Tilde{Y} = M \Tilde{U}$ \textit{characterizes} the implicit models that match both the state and outputs of the baseline model. We then solve a \textit{convex problem} to find another well-posed model, with a desired task in mind, with the matching condition $\Tilde{Y} = M \Tilde{U}$:
\begin{subequations}
\begin{align}
    \min_{M} \;\; &f(M) \\
    \st \quad &Z = AX+BU, \label{eq:sim-state-constraint}\\
    &\hat{Y} = CX+DU, \label{eq:sim-outputs-constraint} \\
    &\|A\|_\infty \le \kappa. \label{eq:sim-wp-constraint}
\end{align}
\end{subequations}
Here, $f$ is an user-designed objective function chosen for a desired task, such as encouraging sparsity, and $\kappa \in (0,1)$ is a hyper-parameter. Note that for a given input matrix $U \in \reals^{p \times m}$, we have generically $U^T U \succ 0$, when $m > p$. The matrix equation $\Tilde{Y} = M \Tilde{U}$ involves $(n+p)m$ scalar equations in $(n+p)(n+q)$ variables, it is thus natural to require that $n > m-p$, which is generally true for over-parameterized models.

The \textit{state-matching constraint} (\ref{eq:sim-state-constraint}) ensures that the implicit model determined by the weight matrices $A, B, C, D$ achieves the same representational power as the baseline model $\mathcal{N}$ by having the same internal state. The \textit{outputs-matching constraint} (\ref{eq:sim-outputs-constraint}) ensures that the model achieves the same predictive performance by obtaining the same predictions as $\mathcal{N}$. Finally, the \textit{well-posedness constraint} (\ref{eq:sim-wp-constraint}) is added to ensure that the well-posedness condition is satisfied. 

For a given baseline layered neural network model, we can always rescale the state matrices $X,Z$ by Theorem (\ref{thm:rescaled}), so that the problem is strictly feasible. Denoting by $W_{\ell}$ the network's matrix corresponding to layer $\ell$, we divide it by the largest max-row-sum norm of the weights among all the layers:
\[
W'_{\ell} = \frac{W_{\ell}}{\gamma \cdot \max_{\ell} \norm{W_{\ell}}_{\infty}}, \; \ell \in [L], \; \gamma > 1,
\]
where $L$ is the total number of layers and $\gamma$ is a scaling factor. The corresponding state matrices $X, Z$ will then be appropriately rescaled after running a single forward pass. 

\subsection{State-driven Training Problem}
SIM is a general training scheme and various kinds of tasks can be achieved by including an appropriately designed objective and setup. We show two such possibilities: one aims for improved sparsity and the other for improved robustness.

\paragraph{Training for sparsity.}
To learn a sparse implicit model, we consider the SIM training problem where we sparsify the weight matrix $M$ by minimizing its cardinality, while satisfying $\Tilde{Y} = M \Tilde{U}$:
\begin{align}
    \label{eq:train-prob-card}
    \min_{M} \quad \norm{M}_0 ~:~ \mbox{(\ref{eq:sim-state-constraint})-(\ref{eq:sim-wp-constraint})}. 
\end{align}
In general, solving the optimization problem (\ref{eq:train-prob-card}) directly is not computationally efficient, and therefore a common alternative is to consider a convex relaxation. We consider the \textit{perspective relaxation} that is a significantly stronger approximation \cite{Frangioni2006PerspectiveCF, atamturk2019rank, Atamtrk2021SparseAS} than the popular $\ell_1$-norm relaxation, and has recently been used for pruning neural networks \cite{PerspectiveFuncPruning}. This leads to the following training problem:
\begin{equation}
    \label{eq:train-prob-perspective}
    \min_{M, t} \;\; \sum_{ij} \frac{M^2_{ij}}{t_{ij}} + \sum_{ij} t_{ij} ~:~ \mbox{(\ref{eq:sim-state-constraint})-(\ref{eq:sim-wp-constraint})}, \; t_{ij} \in [0, 1].
\end{equation}
The perspective terms $\frac{M^2_{ij}}{t_{ij}}$ are typically replaced with auxiliary variables $s_{ij}$ along with rotated cone constraints $M^2_{ij} \le s_{ij} \cdot t_{ij}$ \cite{AAG:conicsch}, leading to a second-order cone problem:
\begin{subequations}
\label{eq:train-prob-perp-socp}
\begin{align}
    \min_{M, t, s} \;\; \alpha \sum_{ij} s_{ij} ~:~ &\mbox{(\ref{eq:sim-state-constraint})-(\ref{eq:sim-wp-constraint})},  \; t_{ij} \in [0, 1], \\
    &M^2_{ij} \le s_{ij} \cdot t_{ij}, \; s_{ij} \ge 0,
\end{align}
\end{subequations}
where $\alpha$ is a hyper-parameter that controls the degree of sparsity. Problem (\ref{eq:train-prob-perp-socp}) can be easily solved with conic quadratic solvers. 

\paragraph{Training for robustness.}
To promote robustness, we consider regularizing the $\ell_1$-norm of the weight matrix $M$. The use of norm-based regularization (\textit{e.g.}  $\ell_2$ or $\ell_1$-norm) for training neural networks has been widely adopted. It has also been shown that there exists an intrinsic relationship between regularizing the $\ell_1$-norm of the weight matrices and their robustness against $\ell_{\infty}$-bounded perturbations \cite{NEURIPS2018_4c5bde74, DBLP:journals/corr/abs-2002-07520}. The set of $\ell_{\infty}$-bounded perturbations yields the worst-case scenario since it includes all other $\ell_p$-bounded perturbations. Controlling the $\ell_1$-norm, therefore, guarantees robustness to $\ell_{\infty}$-perturbations and thereby to all other $\ell_p$-bounded perturbations. Note that we are minimizing the vectorized $\ell_1$-norm of $M$, i.e. $\sum_{ij} |M_{ij}|$, instead of the matrix operator norm. The resulting training problem:
\label{eq:train-prob-l1}
\begin{align}
    \min_{M} \quad \sum_{ij} |M_{ij}| \label{eq:l1-obj} ~:~ \mbox{(\ref{eq:sim-state-constraint})-(\ref{eq:sim-wp-constraint})},
\end{align}
is \textit{convex} and can be solved efficiently by a standard optimization solver.

\paragraph{Relax state and outputs matching.}
We do not have to insist on matching the state and outputs exactly, which allows us to relax the state-matching and output-matching constraints (\ref{eq:sim-state-constraint}) and (\ref{eq:sim-outputs-constraint}) by introducing penalty terms into the objective function:
\begin{subequations}
\label{eq:train-prob-relax}
\begin{align}
    \min_{M} \quad f(M) + \lambda_1 \norm{Z - (AX+BU)}^2_F \\
    + \lambda_2 \norm{\hat{Y} - (CX+DU)}^2_F ~:~ \mbox{(\ref{eq:sim-wp-constraint})}, \; \mathcal{C}, 
\end{align}
\end{subequations}
where $f$ and $\mathcal{C}$ are user-defined objective function and set of constraints on model parameters, respectively, and $\lambda_1$ and $\lambda_2$ are hyper-parameters that control the degree of state- and output-matching.


\paragraph{Parallel training.}
The SIM training problem can be decomposed into a series of parallel, smaller problems, each involving a single row, or a block of rows, if $f$ is decomposable. This is usually the case, including in the sparsity and robustness examples seen before. For a single row $(a^T,b^T)$ of $(A,B)$, and with $z^T$ the corresponding row in $Z$, the problem takes the form of a basis pursuit problem:
\begin{equation} 
    \label{basis-pursuit-ab}
    \mbox{Find vectors } a,b ~:~
    z = \begin{pmatrix} X^T & U^T \end{pmatrix}\begin{pmatrix} a \\ b \end{pmatrix}, \|a\|_1 \le \kappa.
\end{equation}
where $\norm{a}_1 \le \kappa$ is the well-posedness condition since $\norm{A}_{\infty}$ is separable in terms of rows. The problem of finding $C, D$ is independent of that relative to $A, B$ and takes the same form as problem (\ref{basis-pursuit-ab}) without the well-posedness condition:
\begin{equation}
    \label{basis-pursuit-cd}
    \mbox{Find vectors } c,d ~:~ \hat{y} = \begin{pmatrix} X^T & U^T \end{pmatrix}\begin{pmatrix} c \\ d \end{pmatrix}.
\end{equation}
The decomposibility is applicable to the perspective relaxation and the $\ell_1$-norm objective that we consider, with appropriate constraint set $\mathcal{C}$. The parallel SIM training algorithm is summarize in Algorithm \ref{alg:parallel-sim}. More implementation details on parallel training can be found in the appendices.

\begin{algorithm}[!h]
\caption{Parallel \textbf{S}tate-driven \textbf{I}mplicit \textbf{M}odeling (SIM)}
\label{alg:parallel-sim}
    \textbf{Input}: Input feature matrix $U$; A standard neural network or an implicit model $\mathcal{N}: \reals \times \reals \rightarrow \reals \times \reals$; Well-posedness hyper-parameter $\kappa$. \\
    \textbf{Design choices}: Convex minimization objective $f$; Convex constraint set $\mathcal{C}$; Hyper-parameters for $f$. \\
    \textbf{Output}: Weight matrices $A$, $B$, $C$, $D$. \\
    
        \begin{algorithmic}[1] 
        \IF{$\mathcal{N}$ is a standard (layered) neural network}
            \STATE Run a single forward pass on $\mathcal{N}$ with $U$ to obtain outputs $\hat{Y}$, \textit{i.e.,} $\hat{Y} = \mathcal{N}(U)$.
            \STATE Collect all intermediate layers before and after passing through $\phi$.
            \STATE Construct $Z$ and $X$ by stacking all intermediate layers.
        \ELSE
            \STATE Run fix-point iteration until converge for (\ref{eq:eq-eq}) to obtain $X$ and $Z$.
            \STATE Run prediction equation (\ref{eq:pred-rule}) to obatin $\hat{Y}$.
        \ENDIF
        \STATE Put $X, U$ into shared memory.
        \item[] 
        
        \item[\textbf{begin parallel training}]
        \STATE Let $A \gets \mathbf{0} \in \reals^{n \times n}, B \gets \mathbf{0} \in \reals^{n \times p}$ \\ $C \gets \mathbf{0} \in \reals^{q \times n}, D \gets \mathbf{0} \in \reals^{q \times p}$.
        \STATE Distribute rows of $Z$ or $\hat{Y}$ to each processor.
        \FOR{each processor, in parallel}
            \STATE Solve one of the following convex optimization problem: 
                \begin{align*}
                    &\min_{a, b} f(a, b) ~:~ z = \begin{pmatrix} X^T & U^T \end{pmatrix} \begin{pmatrix} a \\ b \end{pmatrix}, \; \norm{a}_1 \le \kappa, \; \mathcal{C} \\
                    &\min_{c, d} f(c, d) ~:~ \hat{y} = \begin{pmatrix} X^T & U^T \end{pmatrix} \begin{pmatrix} c \\ d \end{pmatrix}, \; \mathcal{C}
                \end{align*}
        \ENDFOR
        \STATE Update rows of $A, B$ or $C, D$
        \item[\textbf{end training}]
        \end{algorithmic}
\end{algorithm}

\section{Numerical Experiments}
We demonstrate the capability of SIM on effectively sparsifying and robustifying an implicit model from a given standard (layered) network baseline. We test our method on FashionMNIST \cite{xiao2017fashion} and CIFAR-100 \cite{Krizhevsky2009LearningML} datasets. These experiments were performed on $60$ \textsc{Intel Xeon} processors and solved using \textsc{Mosek} \cite{mosek} optimization solvers. The test set performance is reported. More details on the numerical experiments can be found in the appendices. 

\paragraph{FanshionMNIST.}
We choose a 4-layer fully-connected network of size $784 \times 64 \times 32 \times 16 \times 10$, denoted as $\mathcal{N}_{fc}$, for constructing the state matrices $X, Z$ and the outputs $\hat{Y}$ for FashionMNIST dataset. A Mini batch of size 64 were used for training $\mathcal{N}_{fc}$. The baseline model is trained on a single \textsc{Nvidia Tesla K80} GPU, achieving an $80\%$ test performance.

\paragraph{CIFAR-100.}
For CIFAR-100, we use a ResNet-20 convolutional neural network, denoted as $\mathcal{N}_{res}$. We follow the hyper-parameter settings in \citet{DBLP:journals/corr/abs-1708-04552} with a mini batch of size 128. The model is trained on a single \textsc{Nvidia Titan V} GPU for 200 epochs, with a $75.8\%$ test performance.

\paragraph{Training for sparsity}
In these experiments, we solve the SIM training problem using the perspective relaxation and $\ell_1$-norm objectives with relaxed state and output matching penalties as in problem (\ref{eq:train-prob-relax}), allowing us to control the trade-off between parameter reduction, state-matching, and outputs-matching through hyper-parameters. Throughout the rest of the paper, we use the following hyper-parameters for experiments if not explicitly specified: $\kappa = 0.99$ for well-posedness condition, $\lambda_1 = \lambda_2 = 0.1$ for state-matching and output-matching condition, $\alpha = \num{1e-3}$ for sparsity. To evaluate the performance of parameter reduction, we define sparsity as the percentage of zero parameters of the trained weight matrix $M$ of the total number of non-zero parameters of the baseline model $\mathcal{N}$:
\[
    \mbox{Sparsity (\%)} := \left(1 - \frac{\norm{M}_0}{\norm{\mathcal{N}}_0}\right) \times 100 \cdot
\]

\begin{figure}[!h]
    \centering
    \includegraphics[width=.4\textwidth]{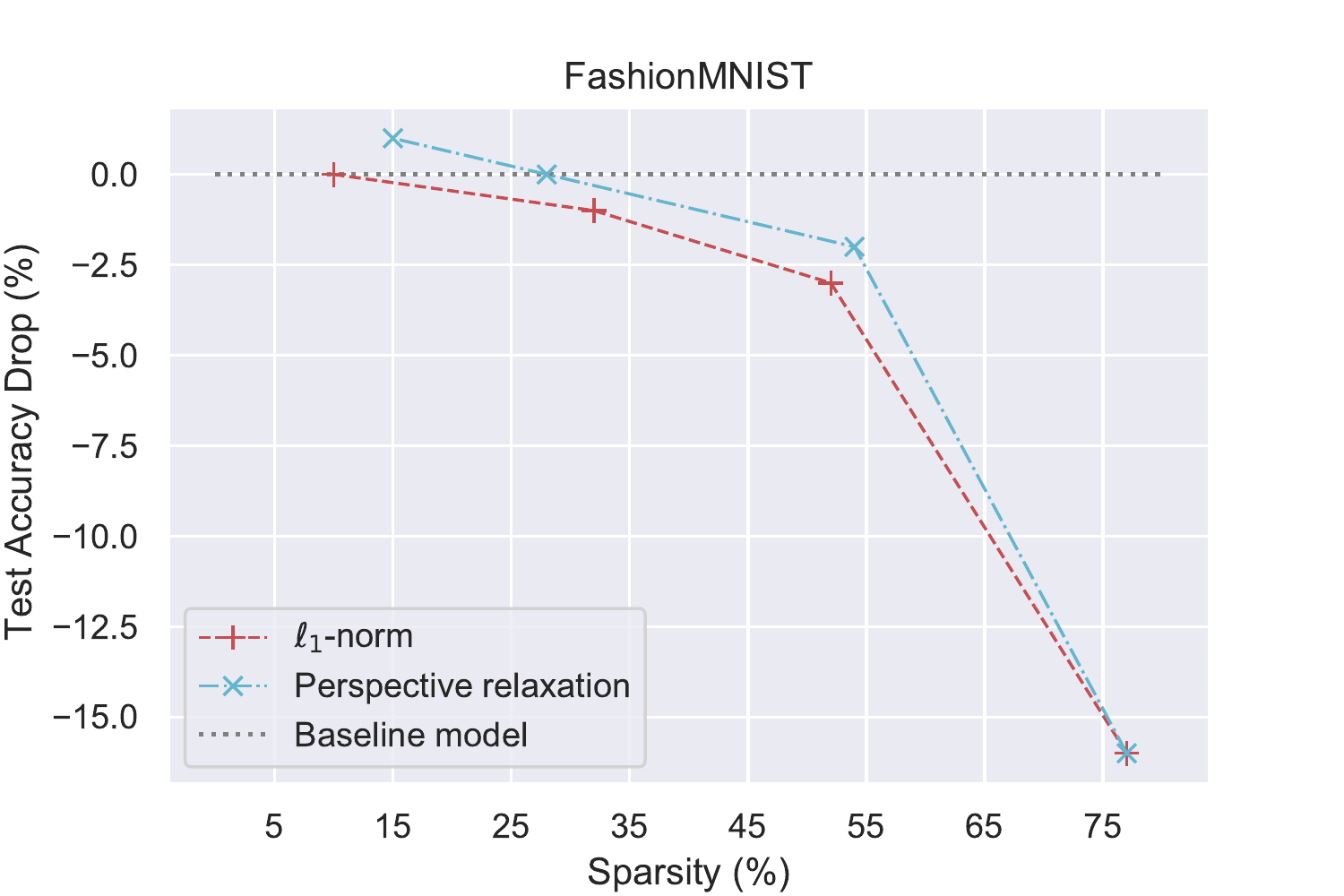}
    \includegraphics[width=.4\textwidth]{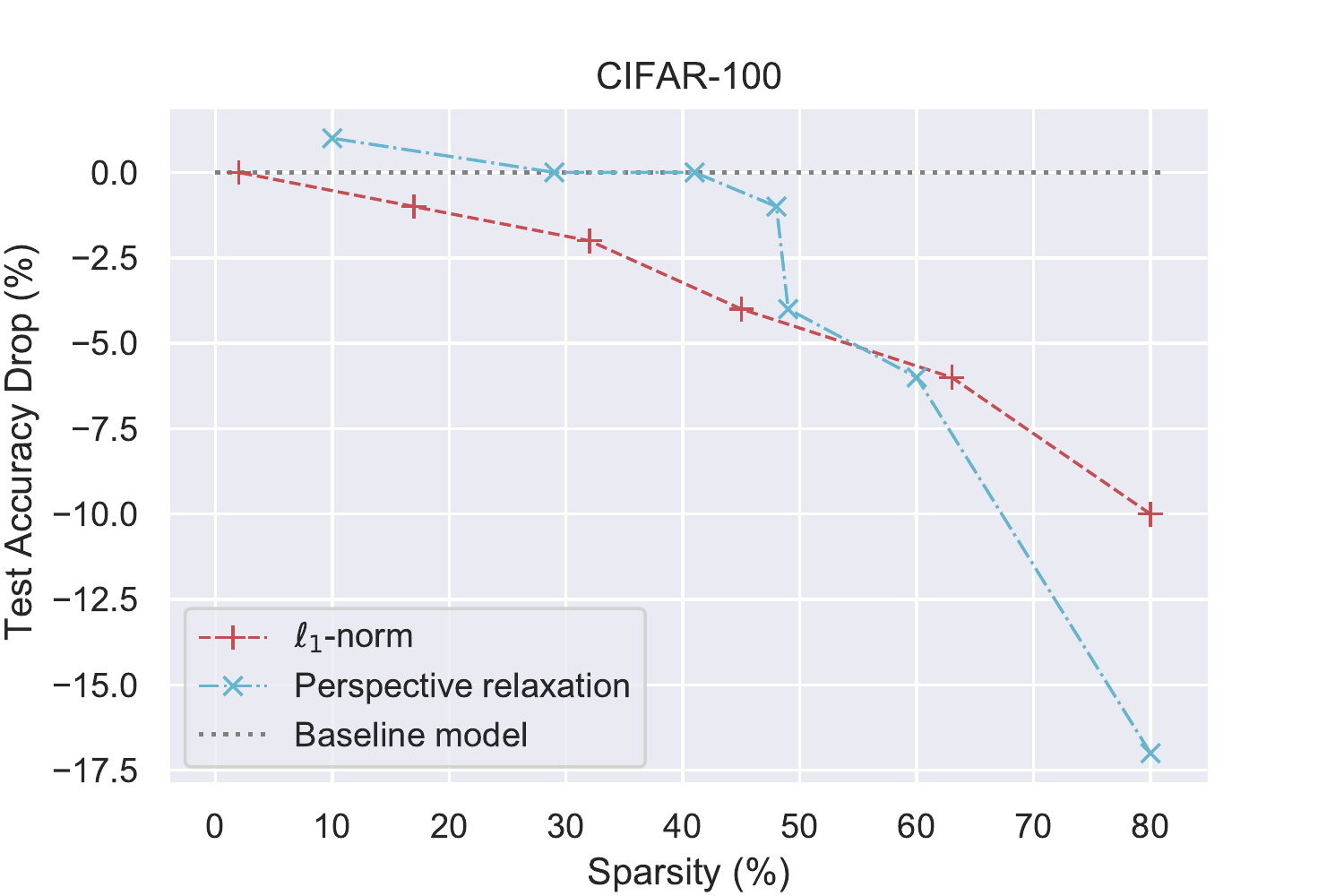}
    \caption{Trade-off curve for sparsity and test accuracy drop for FashionMNIST and CIFAR-100 datasets. Perspective relaxation performs better than $\ell_1$-norm at parameter reduction.}
    \label{fig:sparsity-accuracy}
\end{figure}

Figure \ref{fig:sparsity-accuracy} shows the trade-off curve for sparsity and test accuracy drop for FashionMNIST and CIFAR-100 datasets. The experiments show that using perspective relaxation as objective yields a 28\% reduction and a 41\% reduction of the parameters with no accuracy drop for FashionMNIST and CIFAR-100 datasets respectively. Although $\ell_1$-norm is a more widely used objective for learning sparse models, it is less effective as compared to the perspective relaxation, which is a stronger relaxation for $\ell_0$. Moreover, the perspective relaxation further increases the test performance on both FashionMNIST and CIFAR-100 while reducing 15\% and 10\% of the parameters. 

\begin{figure}[!h]
    \centering
    \includegraphics[width=.4\textwidth]{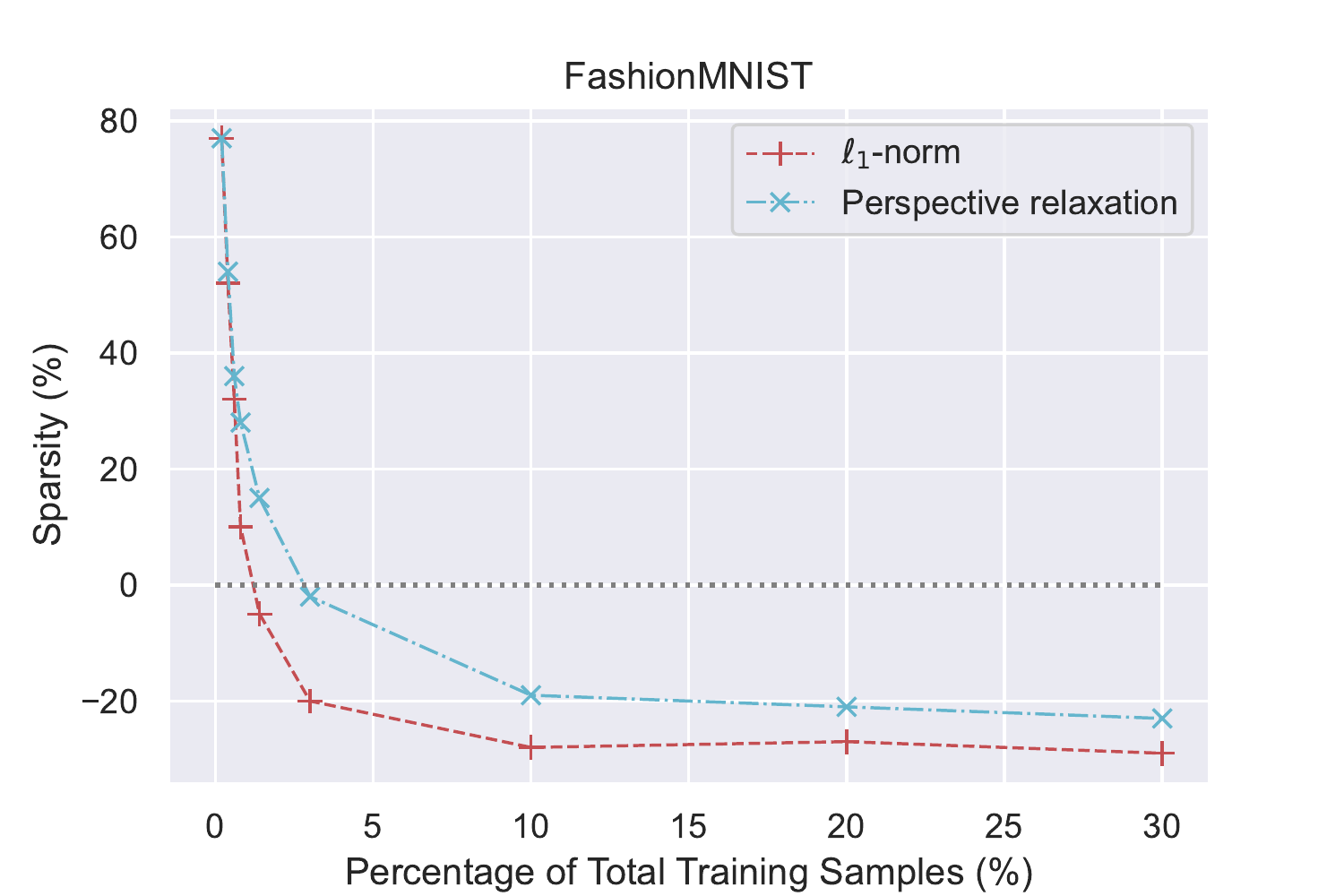}
    \includegraphics[width=.4\textwidth]{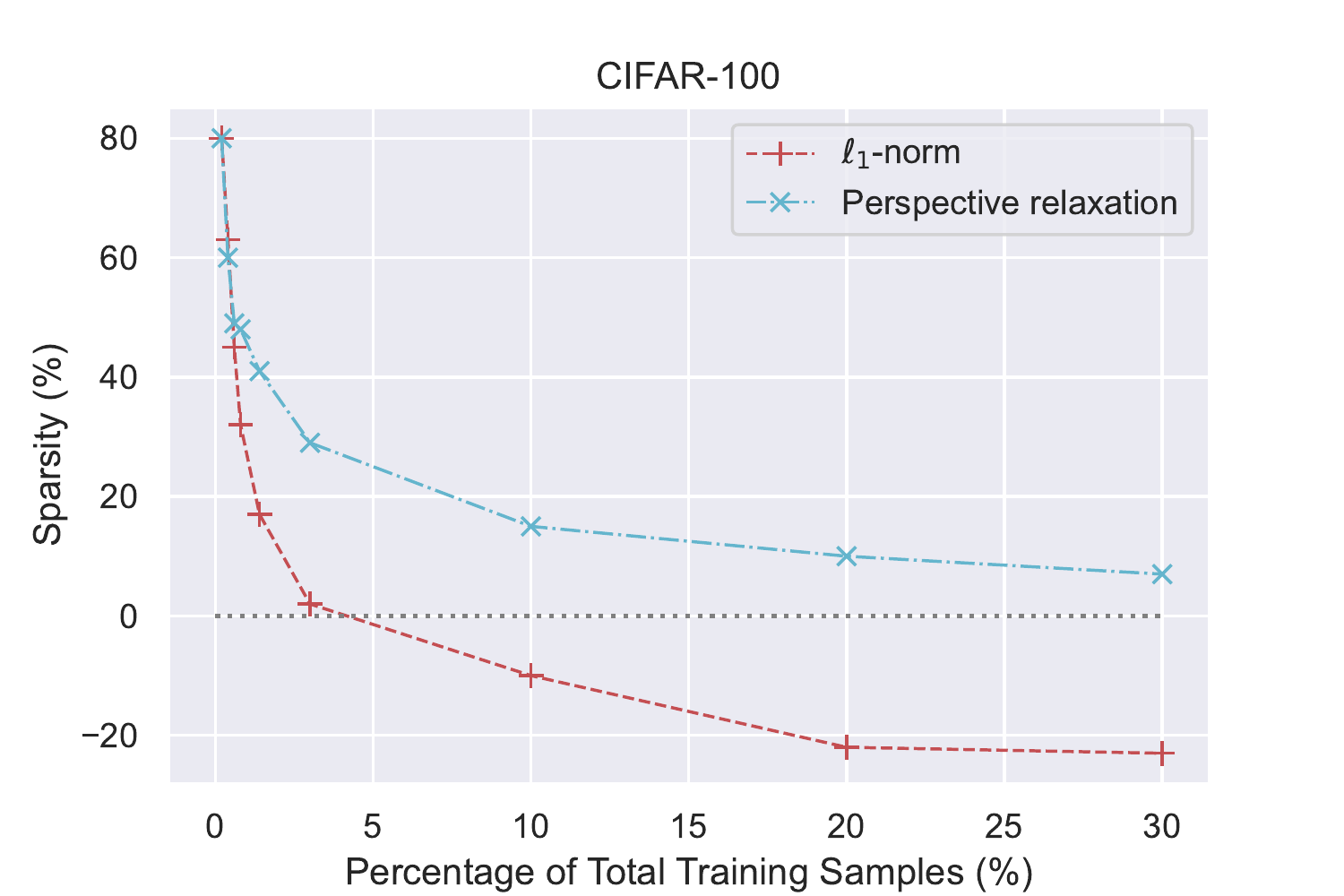}
    \caption{The effective of the number of samples on sparsity for FashionMNIST and CIFAR-100 datasets. Perspective relaxation outperforms the $\ell_1$-norm at parameter reduction for the same number of samples.}
    \label{fig:samples-sparsity}
\end{figure}

In solving problem (\ref{eq:train-prob-relax}), the input matrix $U \in \reals^{p \times m}$ does not have to be the full training dataset. We test how many number of samples are required to sufficiently train a sparse implicit model. Figure \ref{fig:samples-sparsity} shows the effect of the number of samples on sparsity for both datasets. Higher percentage of total training samples means higher $m$ for input matrix $U$. Negative sparsity means that the trained implicit models contain more parameters than the baseline model. We see that for FashionMNIST (50,000 total training samples), we can train a dense model with less than 4\% of the total training data ($\approx$ 2,000 samples) and a sparse model with less than 2\% of the total training data ($\approx$ 1,000 samples). For CIFAR-100 (50,000 total training samples), more data is required as it is a much challenging dataset. Nevertheless, with around 20\% of total training data ($\approx$ 10,000 samples), we are already able to learn a model using perspective relaxation with 10\% fewer parameters. The results indicate that the state matrix $X$ is a high-quality representation that captures a large number of the underlying semantic information, and hence it is sufficient to train a model with significantly fewer training samples. Although the state matrix $X$ is obtained from a standard neural network, we see that in Figure \ref{fig:sparsity-accuracy} we are still able to increase the test performance further with fewer parameters using implicit models. This suggests that implicit models could provide a better representation as compared to a standard layered neural network.

Finally, we compare our method (denoted as SIM) on CIFAR-100 with other parameter reduction methods, including SSS \cite{SSS2018}, SPR \cite{PerspectiveFuncPruning}, and MLA \cite{Hu2019MultiLossAwareCP}. SSS and SPR both formulate the task of pruning as a sparse regularized optimization problem similar to ours, where SSS uses $\ell_1$-relaxation while SPR uses perspective relaxation. MLA considers aligning the semantic information of the intermediate outputs and overall performance of the baseline model and the pruned model by introducing a feature and semantic correlation loss and a classification loss, similar to our state-matching and outputs-matching conditions. SSS and SPR both uses ResNet-20 and MLA uses ResNet-18. We report results of each method as they were reported in the original papers. For SSS, we use the results reported by \citeauthor{PerspectiveFuncPruning}, where the data points are approximated from figures 3(c) of the paper and denoted as P1 and P2. Table \ref{tab:sparsity-compare} shows that SIM achieves less accuracy drop while reducing a larger amount of parameters. With around 30\% of sparsity, SIM has a much lower accuracy drop as compared to MLA and maintains a similar accuracy drop as SPR. With around 45\% of sparsity, SIM outperforms all three methods with a lower accuracy drop and more parameter reduction. It is also likely that additional parameter tuning may lead to more competitive results.

\begin{table}[!h]
    \centering
    \footnotesize
    \caption{\label{tab:sparsity-compare} Comparison with other parameter reduction methods on CIFAR-100.}
    \begin{tabulary}{.5\textwidth}{LCRR}
        \toprule
        \textbf{Method} & \textbf{Setting} & \textbf{Acc. Drop (\%)} & \textbf{Sparsity (\%)} \\
        \midrule
        SSS & P1 Fig. 3(c) & -3.7 & 44.4 \\
        SSS & P2 Fig. 3(c) & -1.3 & 14.8 \\
        SPR & $\lambda=1.0, \alpha=0.1$ & -2.3 & 45.9 \\
        SPR & $\lambda=1.3, \alpha=0.3$ & -0.2 & 31.5 \\
        MLA & ResNet-18 & -3.0 & 50.0 \\
        MLA & ResNet-18 & -2.5 & 30.0 \\
        \midrule
        SIM & Perspective & -1.0 & 48.1 \\
        SIM & Perspective & -0.2 & 29.7 \\
        \bottomrule
    \end{tabulary}
\end{table}

\begin{figure*}[!h]
    \centering
    \includegraphics[width=.4\textwidth]{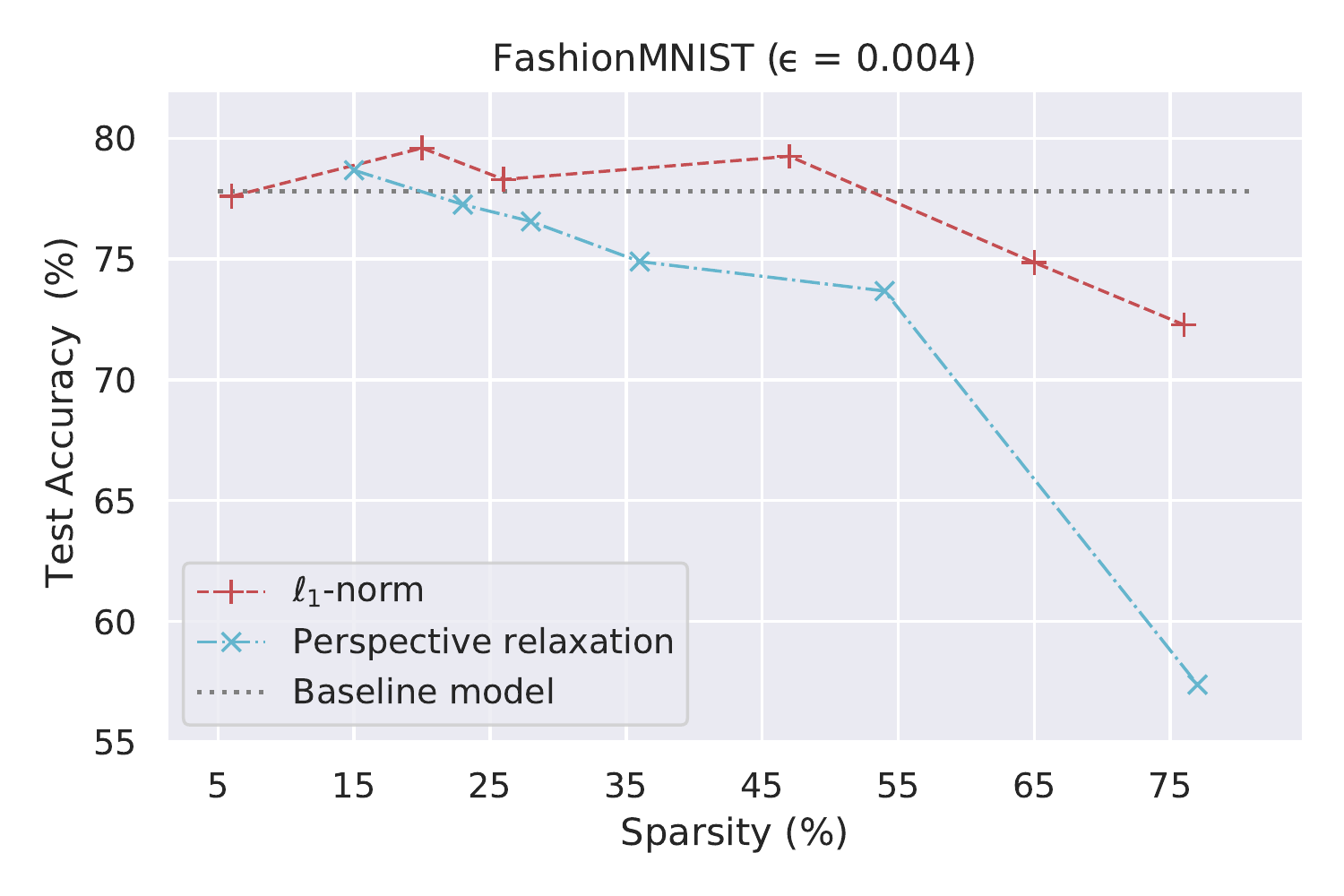}
    \includegraphics[width=.4\textwidth]{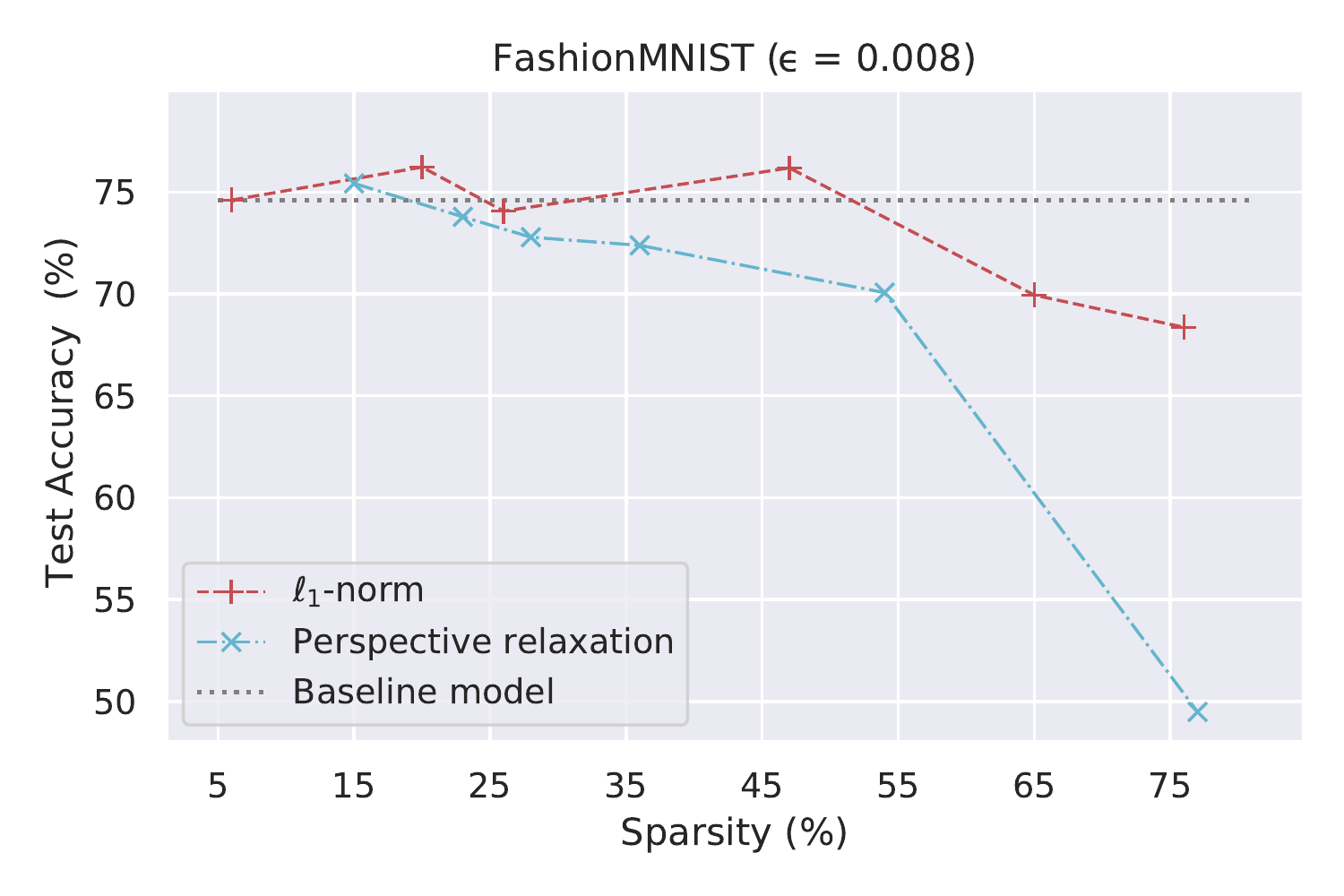}
    \caption{Trade-off curve for robustness with varying weight sparsity. The $\ell_1$-norm is more robust under adversarial attack than the perspective relaxation and the baseline model.}
    \label{fig:robustness}
\end{figure*}

\paragraph{Training for robustness.}
To test for robustness, we perform $\ell_{\infty}$ attacks using the fast gradient sign method (FGSM), presented by \citeauthor{GoodfellowSS14}, on the FashionMNIST dataset to evaluate their adversarial robustness. FGSM generates adversarial examples, $\check{u}$, by taking a step of size $\epsilon$ in the direction of the sign of its gradient taken with respect to the input, $\check{u} = u + \epsilon \cdot \mbox{sgn}(\nabla \mathcal{N}_{fc}(u))$. In our experiments, we set $\epsilon = \frac{1}{255} \approx 0.004$, and $\frac{2}{255} \approx 0.008$. For each batch of the test set, we perturb 50\% of pixels and leave 50\% unperturbed. We evaluate robustness using the prediction accuracy on adversarial examples, $\mathbf{E}_{\check{u}, y} (\mathbf{1}_{y = \mbox{sgn}(f(\check{u}))})$, which measures the ability of a model resisting them. Figure \ref{fig:robustness} illustrates the adversarial robustness with respect to different weight sparsity. In both cases, we observe that $\ell_1$-norm leads to a more robust model as compared to perspective relaxation, and continues to maintain robustness with approximately 45\% fewer parameters. Moreover, the $\ell_1$-norm approach exhibits more robustness than the original baseline network $\mathcal{N}_{fc}$ with higher model sparsity, until the sparsity reaches an over-sparsified threshold that leads to an inevitable capacity degradation.

\paragraph{Visualization of trained models.}

\begin{figure*}[!h]
    \centering
    \includegraphics[width=.85\textwidth]{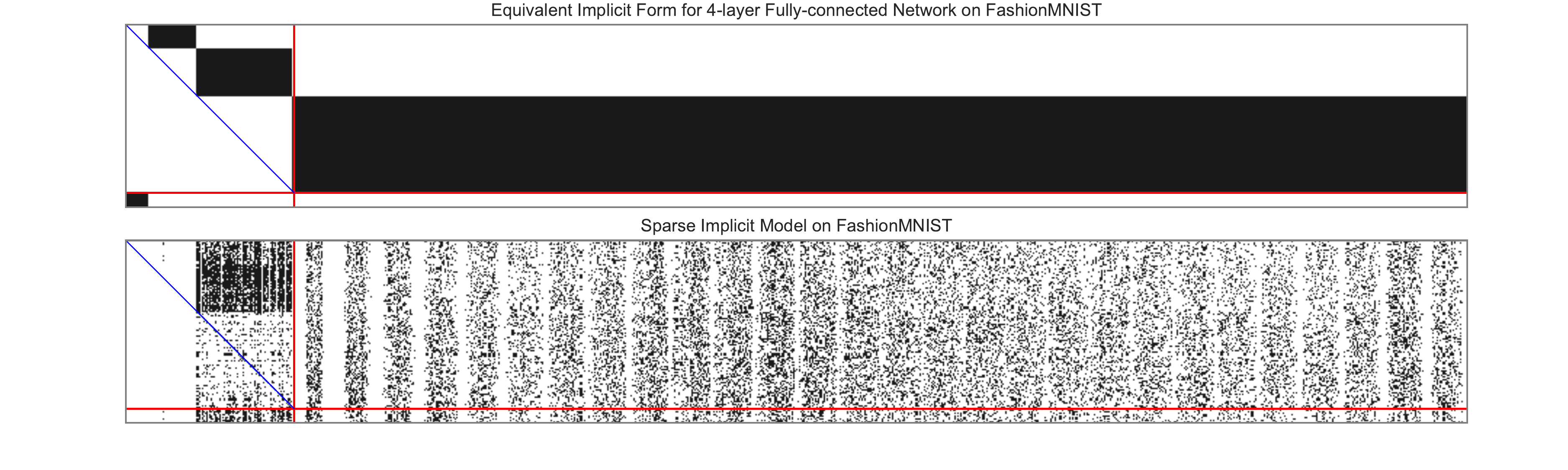}
    \caption{Visualization of the sparsity pattern for the 4-layer fully-connected network $\mathcal{N}_{fc}$ in its equivalent implicit form and a trained sparse implicit model ($\approx$ 30\% less parameters) on FashionMNIST. The weight matrices $A, B, C, D$ correspond to the top left, top right, bottom left, and bottom right location, separated by the read lines. Non-zero parameters are colored in black.}
    \label{fig:fashion-weights-viz}
\end{figure*}

To better understand the properties of the learned model, we plot the weight matrices of the original baseline model versus the sparse implicit model trained on FashionMNIST dataset. Figure \ref{fig:fashion-weights-viz} shows the sparsity pattern for the 4-layer fully-connected network $\mathcal{N}_{fc}$ in its equivalent implicit form and a trained sparse implicit model ($\approx$ 30\% less parameters) on FashionMNIST dataset. We stack the weight matrices in the same way as $M$, where $A, B, C, D$ correspond to the top left, top right, bottom left, and bottom right location respectively. In the visualization, the weight matrices are separated by the red vertical and horizontal lines. The diagonal blue line of the $A$ matrix separates the feed-forward (upper-triangular area) and feedback (lower-triangular area) connections. As shown in Figure \ref{fig:fashion-weights-viz}, a standard layered neural network exhibits a clear hierarchical feed-forward structure, corresponding to upper-diagonal blocks. In contrast, an implicit model allows feedback connections, \ie\ loops in the computational graph. This example illustrates how allowing such loops offers room for marked improvement of the baseline model.

Moreover, we observe that there exists a low-rank structure for the sparse implicit model, where approximately 40\% of the $A$ matrix's dimensions are redundant. This suggests that various dimension reduction methods can be explored to compress the model further; we leave this for future work.

Next, we observe that the $B \in \reals^{112 \times 784}$ matrix of the sparse implicit model also exhibits a sparse pattern where some of the columns are completely zeroed-out. Lastly, for the fully-connected neural network $\mathcal{N}_{fc}$, the large white area on top of the $B_{\mathcal{N}_{fc}}$ matrix indicates there is no connections between the subsequent layers and the input, requiring all low-level information to be effectively captured by the intermediate layers. Nevertheless, allowing the low-level information from the input to pass through directly across the model is desirable and such shortcut connections (or skip connections) have become a common practice in designing deep learning architectures \cite{huang2017densely, he2016deep}. Implicit models allow such skip connections to be learned based on training data without hand engineering the model architecture in advance.   

\section{Conclusion}
In this work, we present the state-driven implicit modeling, a flexible convex optimization scheme for training an implicit model without expensive implicit differentiation, based on fixing the the internal state and outputs from a given baseline model. We describe the convex training problem and  parallel algorithms for training. By introducing an appropriate objective and setup, we demonstrate how state-driven implicit modeling can be applied to train sparse models that are consistently more robust under adversarial attacks. Our results validate the effectiveness of our approach and highlight promising directions for research that bring convex optimization, sparsity, and robustness inducing techniques into implicit modeling.

\bibliography{references}

\newpage
\appendix
\section{Proofs}
\paragraph{Proof of Theorem \ref{thm:pf}}
We first prove the existence of a solution $x \in \reals^{n}$ to the equation $x = \phi(Ax + b)$ if $\lambda_{\text{pf}} < 1$. Since $\phi$ satisfies Assumption (\ref{assum:cone-map}), we have that for $t \ge 1$, the picard iteration 
\[
x_{t+1} = \phi(Ax_t + b), \; x_0 = 0, \; t = 0, 1, \cdots
\]
satisfies
\begin{align*}
    |x_{t+1} - x_t| &= |\phi(Ax_t + b) - \phi(Ax_{t-1} + b)| \\
    &\le |A (x_t - x_{t-1})| \le |A| |x_t - x_{t-1}| \\
    &\le |A|^t |x_1 - x_0|.
\end{align*}
Hence, for every $t, \tau \ge 1$, we have 
\begin{align*}
    & |x_{t+\tau} - x_t| = \left|\sum_{i=t+1}^{t+\tau} (x_i - x_{i-1})\right| \le \sum_{k=t}^{t+\tau} |A|^k |x_1 - x_0| \\
    & \le |A|^t \sum_{k=0}^{\tau} |A|^k |x_1 - x_0| \le |A|^t \sum_{k=0}^{\infty} |A|^k |x_1 - x_0| \\
    & = |A|^t (I - |A|)^{-1} |x_1 - x_0|.
\end{align*}
The inverse of $I - |A|$ exists as $\lambda_{\text{pf}}(|A|) < 1$. Since $\lim_{t \to \infty} |A|^t = 0$, we have 
\[
0 \le \lim_{t \to \infty}|x_{t+\tau} - x_t| \le \lim_{t \to \infty} |A|^t (I - |A|)^{-1} |x_1 - x_0| = 0.
\]
We obtain that $x_t$ is a Cauchy sequence, and thus the sequence converges to some limit point, $x_\infty$, which by continuity of $\phi$ can be obtained by $x_\infty = \phi(A x_\infty +b)$, thus establishes the existence of a solution to $x = \phi(Ax + b)$.

For uniqueness, consider two solutions $x_a, x_b \in \reals^n_{+}$ to the equation, the following inequality holds,
\[
0 \le |x_a - x_b| \le |A| |x_a - x_b| \le |A|^k |x_a - x_b|.
\]
As $k \to \infty$, we have that $|A|^k \to 0$, and it follows that $x_a = x_b$, which establishes the unicity of the solution.

\paragraph{Proof of Theorem \ref{thm:rescaled}}
Consider a neural network $\mathcal{N}$ in its equivalent implicit form $(A_\mathcal{N}, B_\mathcal{N}, C_\mathcal{N}, D_\mathcal{N}, \phi)$, since the matrix $|A_{\mathcal{N}}|$ is strictly upper triangular, all of its eigenvalues are zeros, automatically satisfying the PF sufficient condition for well-posedness. From the Collatz-Wielandt formula \cite{10.5555/343374}, the PF eigenvalue of a well-posed implicit model can be represented as 
\[
\lambda_{\text{pf}}(|A|) = \inf_{s > 0} \norm{\diag{s} |A| \diag{s}^{-1}}_{\infty}.
\]
The scaling factor $s$ such that $\norm{\diag{s} |A| \diag{s}^{-1}}_{\infty} < 1$ can be obtained by solving
\[
s_i = 1 + \sum_{j=i+1}^n |A_{i, j}| s_j, \; i \in [n],
\]
which can then be solved by backward substitution. The new model matrices $(A', B', C', D', \phi)$, are obtained by
\[
\renewcommand\arraystretch{1.3}
\left( \begin{array}{c|c} A' & B' \\ \hline C' & D' \end{array} \right)=  \left( \begin{array}{c|c} SAS^{-1} & SB \\ \hline CS^{-1} & D \end{array} \right)
\]
where $S = \diag{s}$, with $s > 0$ a PF eigenvalue of $|A|$. More generally, provided that $\lambda_{\text{pf}}(|A|) < 1$, we simply set $s = (I - |A|)^{-1}\mathbf{1}$, which can be obtained as the limit point of fixed-point iterations.

\section{More on Parallel Training}
\label{app:parallel-train}
\paragraph{Data structure.}
Fitting all the weight matrices into memory requires a substantial amount of storage space. However, we can leverage the high-sparsity property of the problem to reduce the memory consumption when storing the weight matrices. In the high-sparsity regime, schemes known from high-performance computing such as compressed sparse row (CSR) and compressed sparse column (CSC) can store indices of matrices, respectively. Since in this problem, we operate in a row-wise fashion, we choose to store the weight matrices in CSR format. CSR represents the indices in an $n = n_r \times n_c$ matrix using row and column index arrays. The row array is of length $n_r$ and store the offsets of each row in the value array in $\lceil \log_2 m \rceil$ bits, where $m$ is the number of non-zero elements. The column array is of length $m$ and stores the column indices of each value in $\lceil \log_2 n_c \rceil$ bits. The total storage space required is therefore $n_r \times \lceil \log_2 m \rceil + m \times \lceil \log_2 n_c \rceil$.

\paragraph{Multiprocessing.}
Given state matrices from a neural network, the basis pursuit problem of (\ref{basis-pursuit-ab}) and (\ref{basis-pursuit-cd}) can be paralleled, each involving a single or a block of rows. Each block is trained independently by a child processor with an auxiliary objective, and returns the solutions back to the main processor. We implement our parallel training algorithm with the \textsc{multiprocessing} package using Python. The \textsc{multiprocessing} package\footnote{\url{https://docs.python.org/3/library/multiprocessing.html}} supports spawning processes and offers both local and remote concurrency. In Python, its Global Interpreter Lock (GIL) only allows one thread to be run at a time under the interpreter, which means we are unable to leverage the benefit of multi-threading. However, with multiprocessing, each process has its own interpreter and the instructions are executed by its own interpreter, which allows multiple processes to be run in parallel, side-stepping the GIL by using sub-processes instead of threads. In \textsc{multiprocessing}, a process is a program loaded into memory to run and does not share its memory with other processes. The decomposability of the training problem can be viewed as \textit{data parallelism} where the execution of a function, i.e. solving the convex optimization problem, is parallelized, and the input values are distributed across processes. We use the \verb|Pool| object to offer a means of defining a function in a module so that child processes can each import the module and execute it independently. 

\paragraph{Memory sharing.} In \textsc{multiprocessing}, data in the arguments are pickled and passed to the child processors by default. In the basis pursuit problem, the state matrix $X$ and the input data matrix $U$ remain unchanged during task execution across all the processors, and thus only need read-only access to $X$ and $U$. Passing $X$ and $U$ to each processor whenever a new task is scheduled consumes a significant amount of memory space and increases the communication time. As a result, instead of treating them as data input to the function, we put $X$ and $U$ into a shared memory, providing direct access of the shared resources across processes.

\paragraph{Ray.}
We also implement our parallel algorithm using \textsc{Ray}\footnote{\url{https://www.ray.io/}}, an open-source and general-purpose distributed compute framework for machine learning and deep learning applications. By transforming the execution of the convex training problems into ray \verb|actors|, we are able to distribute the input values to multiple ray actors to run on multiple ray nodes. Similar to the memory sharing in the multiprocessing approach, we use \verb|ray.put()| to save objects into the ray object store, saving memory bandwidth by only passing the object ids around. We run our experiments on the Cori clusters\footnote{\url{https://docs.nersc.gov/systems/cori/}} hosted by National Energy Research Scientific Computing (NERSC) Center and use the \textsc{slurm-ray-cluster} scripts\footnote{\url{https://github.com/NERSC/slurm-ray-cluster}} for running multi-nodes.

\begin{figure}[!h]
    \centering
    \includegraphics[width=.45\textwidth]{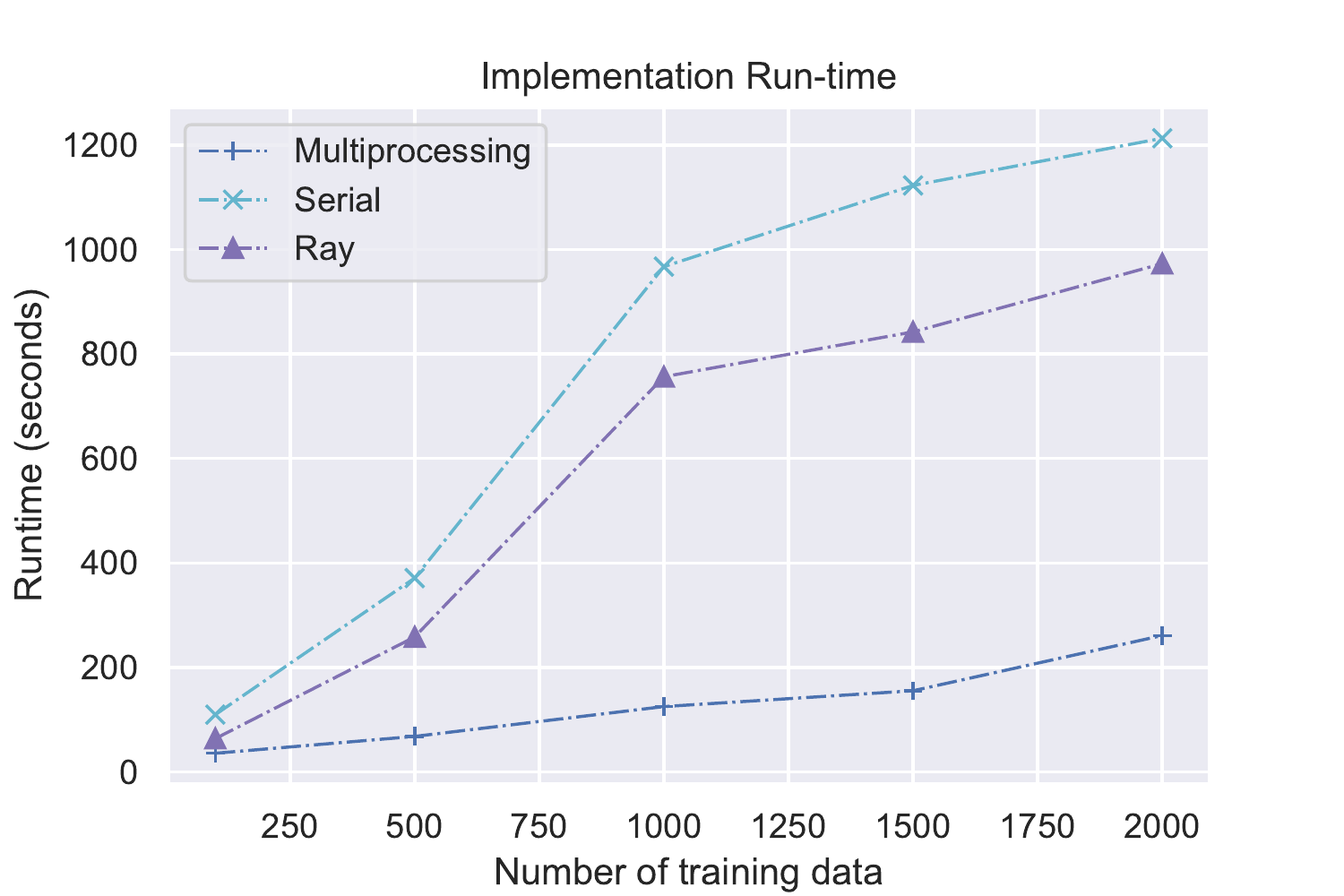}
    \caption{Performance benchmark for serial, multiprocessing (parallel), and Ray (parallel) implementation on FashionMNIST dataset using 8 processors.}
    \label{fig:runtime}
\end{figure}

\paragraph{Performance benchmark.}
Figure \ref{fig:runtime} show the run-time for our serial and parallel implementation using both \textsc{multiprocessing} and \textsc{Ray}. We observe that \textsc{multiprocessing} provides the best speedup as compared to \textsc{Ray}. We hypothesize that since \textsc{Ray} is a general-purpose distributed compute framework, it contains more overhead than solving the training problem directly using \textsc{multiprocessing}.

\section{More on Numerical Experiments}
\label{app:experiments}
Table \ref{tab:perspective-settings} and Table \ref{tab:l1-settings} shows the number of training samples, hyper-parameters, and adversarial test accuracy for perspective relaxation and $\ell_1$-norm objective functions with state and outputs matching penalties as in problem (\ref{eq:train-prob-relax}). For perceptive relaxation, we solve the following problem:
\begin{subequations}
\begin{align}
    \min_{M, t, s} \;\; &\alpha \sum_{ij} s_{ij} + \lambda_1 \norm{Z - (AX+BU)}^2_F \\
    &+ \lambda_2 \norm{\hat{Y} - (CX+DU)}^2_F \\
    \st \quad &\mbox{(\ref{eq:sim-wp-constraint})},  \; t_{ij} \in [0, 1], M^2_{ij} \le s_{ij} \cdot t_{ij}, \; s_{ij} \ge 0.
\end{align}
\end{subequations}
For the $\ell_1$-norm problem, we solve the following problem:
\begin{subequations}
\begin{align}
    \min_{M} \quad & \beta \sum_{ij} |M_{ij}| + \lambda_1 \norm{Z - (AX+BU)}^2_F \\
    &+ \lambda_2 \norm{\hat{Y} - (CX+DU)}^2_F ~:~ \mbox{(\ref{eq:sim-wp-constraint})},
\end{align}
\end{subequations}
where $\beta$ controls the degree of regularizing for robustness.

\begin{table}[!h]
    \centering
    \footnotesize
    \caption{\label{tab:perspective-settings} Experimental settings for perspective relaxation on Fashion-MNIST.}
    \begin{tabulary}{.5\textwidth}{C CC CC CC}
    \toprule
    & & & & & \multicolumn{2}{c}{\textbf{Test Acc. (\%)}} \\
    \cmidrule{6-7}
    \textbf{\# Train Samples} & \textbf{Sparsity (\%)} & \textbf{$\lambda_{1}$} & \textbf{$\lambda_{2}$} &  \textbf{$\alpha$} & \textbf{$\epsilon = 0.004$} & \textbf{$\epsilon = 0.008$}\\
    \midrule
    700 & 15 & 0.1 & 0.1 & 0.01 & 78.7 & 75.4 \\
    500 & 23 & 0.1 & 0.1 & 0.01 & 77.3 & 73.8 \\
    400 & 28 & 0.1 & 0.1 & 0.01 & 76.6 & 72.8 \\
    300 & 36 & 0.1 & 0.1 & 0.01 & 74.9 & 72.4 \\
    200 & 54 & 0.1 & 0.1 & 0.01 & 73.7 & 70.1 \\
    100 & 77 & 0.1 & 0.1 & 0.01 & 57.2 & 49.5 \\
    \bottomrule
    \end{tabulary}
\end{table}

\begin{table}[!h]
    \centering
    \footnotesize
    \caption{\label{tab:l1-settings} Experimental settings for $\ell_1$-norm on Fashion-MNIST.}
    \begin{tabulary}{.5\textwidth}{C CC CC CC}
    \toprule
    & & & & & \multicolumn{2}{c}{\textbf{Test Acc. (\%)}} \\
    \cmidrule{6-7}
    \textbf{\# Train Samples} & \textbf{Sparsity (\%)} & \textbf{$\lambda_{1}$} & \textbf{$\lambda_{2}$} &  \textbf{$\beta$} & \textbf{$\epsilon = 0.004$} & \textbf{$\epsilon = 0.008$}\\
    \midrule
    600 & 20 & 0.1 &  0.1 &  0.001 & 79.6 & 76.2 \\
    1000 & 47 & 0.01 &  0.01 &  0.001 & 79.3 & 76.2 \\
    500 & 26 & 0.01 & 0.01 &  0.01 & 78.3 & 75.0 \\
    2000 & 6 & 0.01 &  0.01 &  0.01 & 77.6 & 74.6 \\
    900 & 65 &  0.01 &  0.01 &  0.001 & 74.9 & 69.9 \\
    400 & 76 & 0.01 &  0.01 &  0.001 & 72.3 & 68.4 \\
    \bottomrule
    \end{tabulary}
\end{table}

\end{document}